\documentclass[10pt,twocolumn,letterpaper]{article}

\usepackage{ijcb}
\usepackage{times}
\usepackage{epsfig}
\usepackage{graphicx}
\usepackage{amssymb}

\usepackage{booktabs}

\usepackage{hyperref}
\usepackage[table]{xcolor}

\usepackage[accsupp]{axessibility} 

\usepackage{algorithm}
\usepackage{algpseudocode}
\usepackage{amsmath,bm}
\usepackage{amsfonts}
\usepackage{multirow}
\usepackage{xcolor}

\definecolor{Gray}{gray}{0.94} 

\ijcbfinalcopy 

\begin{document}

\title{SynthDistill: Face Recognition with Knowledge Distillation from Synthetic Data}

\author{\vspace{4pt}Hatef Otroshi Shahreza$^{1,2}$, Anjith George$^{1}$, 
			S\'{e}bastien Marcel$^{1,3}$\\
   $^{1}$Idiap Research Institute, Martigny, Switzerland\\
   $^{2}$\'{E}cole Polytechnique F\'{e}d\'{e}rale de Lausanne (EPFL), Lausanne, Switzerland  \\
   $^{3}$Universit\'{e} de Lausanne (UNIL), Lausanne, Switzerland\\
   {\tt\small \{hatef.otroshi,anjith.george,sebastien.marcel\}@idiap.ch}
}

\maketitle
\thispagestyle{empty}

\begin{abstract}
State-of-the-art face recognition networks are often computationally expensive and cannot be used for mobile applications. 
Training lightweight face recognition models also requires large identity-labeled datasets. Meanwhile, there are privacy and ethical concerns with collecting and using large face recognition datasets. 
While generating synthetic datasets for training  face recognition models is an alternative option, it is challenging to generate synthetic data with sufficient intra-class variations. In addition, there is still a considerable gap between the performance of models trained on real and synthetic data.
In this paper, we propose a new framework (named SynthDistill)  to train lightweight face recognition models by distilling the knowledge of a pretrained teacher face recognition model using synthetic data. 
We use a pretrained face generator network to generate synthetic face images and use the synthesized images to learn a lightweight student network.  We use synthetic face images without identity labels, mitigating the problems in the intra-class variation generation of synthetic datasets. 
Instead, we propose a novel dynamic sampling strategy from the intermediate latent space of the face generator network to include new variations of the challenging images while further exploring new face images in the training batch.  
The results on five different face recognition datasets demonstrate the superiority of our lightweight model compared to models trained on previous synthetic datasets, achieving a verification accuracy of 99.52\% on the LFW dataset with a lightweight network. The results also show that our proposed framework significantly reduces the gap between training with real and synthetic data.
The source code for replicating the experiments is publicly released.
\end{abstract}

\section{Introduction}

Recent advancements in face recognition systems have been driven by deep neural networks trained on large-scale datasets, leading to remarkable progress in accuracy~\cite{deng2019arcface,kim2022adaface}.
However, the state-of-the-art face recognition networks are often computationally heavy and the deployment of these networks on edge devices poses practical challenges. Nevertheless, it is possible to develop efficient networks from these large models that achieve comparable accuracy with significantly reduced computational load, making them suitable for edge device deployment.

\begin{figure}[t!]
    \centering
    \includegraphics[width=1\columnwidth]{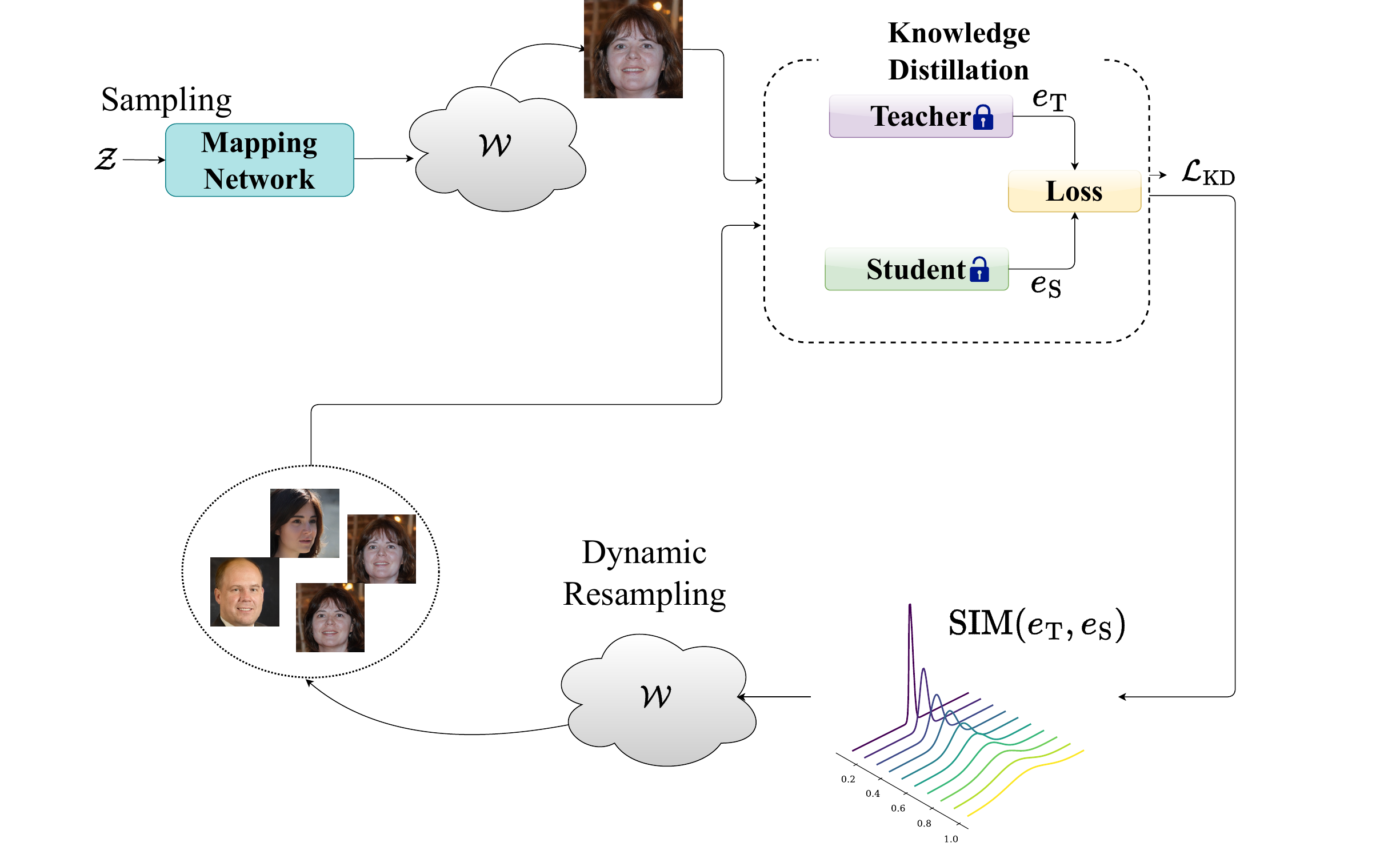}
    \caption{Schematic showing the proposed approach (SynthDistill). Latent space of StyleGAN is first sampled from $\mathcal{Z}$ space, and then dynamically re-sampled from $\mathcal{W}$ space based on teacher-student agreement. This dynamic re-sampling leads to the generation of challenging samples that facilitate efficient learning.
    } \label{fig:GA}
\end{figure}

One strategy is training lightweight and efficient networks on the large-scale face recognition datasets~\cite{alansari2023ghostfacenets,boutros2021mixfacenets,boutros2022pocketnet,george2023edgeface,kolf2023efar_arxiv_from_scholar}. However, training an efficient face recognition model using large-scale face recognition datasets requires access to such a dataset. 
Nonetheless, large-scale face recognition datasets, such as VGGFace2~\cite{cao2018vggface2}, MS-Celeb~\cite{guo2016ms}, WebFace~\cite{zhu2021webface260m}, etc., were collected by crawling images from the Internet, thus raising legal, ethical, and privacy concerns~\cite{boutros2023synthetic}. 
To address such concerns, recently several works proposed generating  synthetic face datasets and use the synthetic face images for training face recognition models~\cite{bae2023digiface,boutros2022sface,kim2023dcface}. However, generating synthetic face datasets with sufficient inter-class and intra-class variations is still a challenging problem. Our experimental results also show that there is still a large gap in the recognition performance when training a lightweight face recognition model on real data and existing synthetic face datasets.

Another strategy to train a lightweight face recognition model is to transfer the knowledge of a model trained on a large dataset to a lightweight network through knowledge distillation \cite{hinton2015distilling}. 
Notwithstanding, the knowledge distillation from a teacher model often requires access to the original or another large-scale real dataset. 
Meanwhile, access to a real dataset for knowledge distillation may not always be feasible due to the size of the datasets. 
Even if there is access to real large-scale dataset, there remain ethical and legal concerns of using large-scale face recognition datasets crawled from internet. 
In this work, we propose a new framework to distill  the knowledge of a pretrained teacher using synthetic face images without identity labels, and thus mitigating the need for real identity-labeled data during the distillation phase. We propose dynamic sampling from the intermediate latent space of a StyleGAN to generate new images and enhance training.

In contrast to previous approaches that rely on static generation of synthetic face datasets ~\cite{bae2023digiface,boutros2022sface,kim2023dcface}  and then using the generated dataset for training the FR model, 
we combine these two steps with an online-generation of synthetic images and training the lightweight network in the image generation loop within a knowledge distillation based framework. This avoids the requirements of hard identity labels for the generated images, and further assists the generation network to produce challenging samples  though a feedback mechanism while exploring more image variations, thus enabling the training of more robust models. 
In addition, compared to previous works for the training of face recognition models on synthetic datasets, our proposed knowledge distillation framework does not require identity labels in the training, simplifying the process of  generating synthetic face images. 
We should also note that previous synthetic datasets still used a face recognition model in the dataset generation pipeline. 

In our case, we also employ a pre-trained face recognition model in our pipeline, but with the role as a teacher. However, instead of generating a static synthetic dataset with identity labels, we dynamically create synthetic face images during the knowledge distillation process. This novel approach allows us to frame our knowledge distillation as a label-free training paradigm, utilizing synthetic data to effectively train lightweight face recognition models.

It is noteworthy that we do not need access to the complete whitebox knowledge of the teacher network in our proposed knowledge distillation approach, and thus our method can also be used in case of a blackbox access to the teacher model that can used to generate the embeddings, given the embeddings are available. 
We adapt the TinyNet~\cite{han2020model} architecture and train lightweight face recognition models (called \textit{TinyFaR}) in our  knowledge distillation approach.  
We provide an extensive experimental evaluation on five different face recognition benchmarking datasets, 
including LFW \cite{huang2008labeled}, CA-LFW \cite{zheng2017cross}, CP-LFW \cite{zheng2018cross}, CFP-FP \cite{sengupta2016frontal} and AgeDB-30 \cite{moschoglou2017agedb}. 
Our experimental results demonstrate 
the effectiveness of our approach in achieving efficient face recognition systems with reduced computational requirements, while avoiding the use of real data for knowledge distillation. This opens new possibilities for developing privacy-aware and resource-efficient face recognition models suitable for edge devices. Fig.~\ref{fig:GA}
 illustrates the general block diagram of our proposed knowledge distillation framework with dynamic sampling.

The main contributions of this work are listed below:
\begin{itemize}
    \item We propose a novel framework to train a lightweight face recognition model using knowledge distillation. 
    The proposed knowledge distillation framework is based on synthetic face images and does not require real training data. In addition, we do not need identity-labeled training data in our knowledge distillation framework, mitigating problems in generating synthetic face recognition datasets. 
     \item Our proposed knowledge distillation framework is based on a dynamic sampling of difficult samples during training to enhance the training. Dynamic sampling helps the student network to simultaneously learn on new images (i.e., increase generalization), while focusing on difficult samples. Therefore, the training images are synthesized online and during the distillation process.
    \item We provide extensive experimental results on different face recognition datasets, showing superior recognition accuracy for lightweight face recognition models trained in our framework compared to training lightweight face recognition from scratch using other synthetic datasets. 
\end{itemize}

The remainder of the paper is organized as follows.
In Section~\ref{sec:related-works} we review the related works in the literature. 
We describe our proposed framework for  knowledge distillation with synthetic data using dynamic latent sampling in Section~\ref{sec:proposed-method}. We report our experimental results in Section~\ref{sec:experiments} and also discuss our results in Section~\ref{sec:discusssion}. Finally, the paper is concluded in Section~\ref{sec:conclusion}.

\section{Related works}\label{sec:related-works}
In this section, we discuss the relevant literature on synthetic datasets, light-weight face recognition networks, and knowledge distillation in the context of face recognition.

\subsection{Synthetic Datasets}

Several works have explored the generation of synthetic datasets for training face recognition. It is worth noting that many large-scale datasets are typically collected through web-crawling without explicit informed consent. By leveraging synthetic datasets, it becomes possible to mitigate concerns regarding the privacy of individuals while also potentially addressing issues such as bias~\cite{jain2023zero,sevastopolsky2022boost}. These synthetic datasets are often generated using variations of StyleGAN, 3D models, and diffusion models.

Several prior works, including FaceID-GAN \cite{shen2018faceid}, identity-preserving face images \cite{bao2018towards} \cite{yin2017towards}, have employed synthesis techniques to generate facial images. Notably, FF-GAN \cite{yin2017towards} (e.g., 3DMM \cite{blanz1999morphable}) and DiscoFaceGAN \cite{deng2020disentangled} leverages 3D priors. In \cite{qiu2021synface}, authors proposed an approach called SynFace which incorporates the use of identity mixup (IM) and domain mixup (DM) techniques to address the performance gap. They use a small portion of labeled real data in the training process to reduce the domain gap between real and synthetic data to improve the performance. Additionally, the controllable face synthesis model provides a convenient means to manipulate various aspects of synthetic face generation, such as pose, expression, illumination, the number of identities, and samples per identity. Boutros et al. \cite{boutros2022sface}, presented a method to generate synthetic data using a class conditional generative adversarial network. The authors trained the StyleGAN2-ADA model \cite{karras2020training} on the CASIA-WebFace \cite{yi2014learning} datasets, using identities as class labels. They have conducted experiments using the generated SFace dataset to show its utility in training face recognition models. 
Bae et al. \cite{bae2023digiface}, introduced a large-scale synthetic dataset for face recognition named DigiFace-1M. This dataset was created by utilizing a computer graphics pipeline to render digital faces. Each identity within the dataset is generated by incorporating randomized variations in facial geometry, texture, and hairstyle. The rendered faces exhibit diverse attributes such as different poses, expressions, hair color, hair thickness, and density, as well as accessories. Through the implementation of aggressive data augmentation techniques, they reduced the domain gap between the generated images and real face images leading to gains in face recognition performance. In \cite{kim2023dcface}, authors proposed a Dual Condition Face Generator (DCFace) utilizing a diffusion model. This approach incorporates a novel Patch-wise style extractor and Time-step dependent ID loss, enabling DCFace to consistently generate face images depicting the same individual in different styles, while maintaining precise control over the process. 

Despite the advantages of synthetic data in terms of privacy and consent, the performance of face recognition models trained on these datasets falls short when compared to models trained on real data. This severely limits real-world usage of models trained on synthetic datasets.  To address these challenges, we propose a novel strategy for training face recognition models using synthetic data within an kowledge distillation framework. Our method generates data online dynamically and eliminates the need for real data during the distillation phase.

\begin{figure*}[t]
    \centering
    \includegraphics[width=0.99\linewidth]{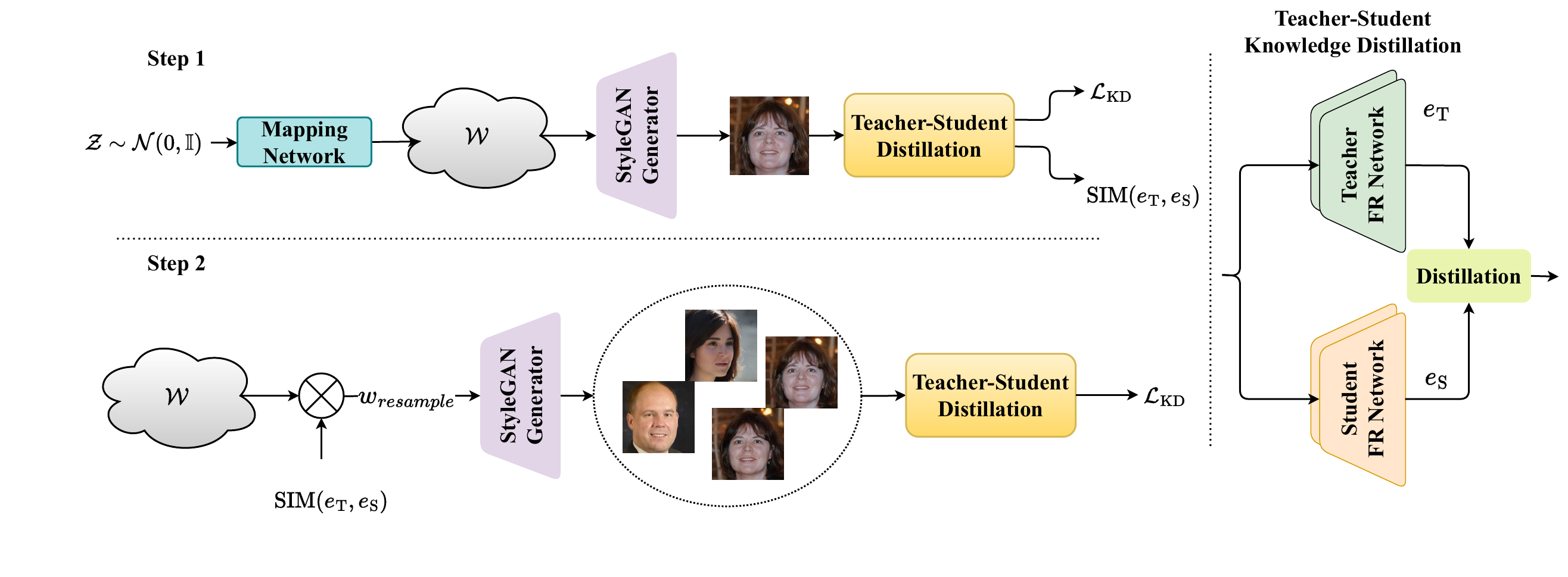}
    \caption{Schematic showing the proposed approach (SynthDistill). In step 1, $\mathcal{Z}$ space of the StyleGAN is sampled to generate face images. In step 2, the $\mathcal{W}$ space is re-sampled based on the teacher-student agreement to generate more challenging samples. The student model is is updated based on the distillation loss $\mathcal{L}_{\text{KD}}$, all the other network blocks remains frozen.
    } \label{fig:framework}
  \end{figure*}

\subsection{Efficient Face Recognition}

As edge computing gained prevalence, there is an increased focus on developing lightweight face recognition models without compromising accuracy. In the initial phase of efficient model development, Wu et al. introduced LightCNN, a lightweight architecture \cite{wu2018light}. MobileNets \cite{howard2017mobilenets, sandler2018mobilenetv2} employed depth-wise separable convolutions to improve the performance. Building upon the MobileNet architecture, MobileFaceNets were designed for real-time face verification tasks \cite{chen2018mobilefacenets}. The concept of MixConv, which incorporates multiple kernel sizes in a single convolution, was used to develop MixFaceNet networks for lightweight face recognition \cite{tan2019mixconv, boutros2021mixfacenets}. Inspired by ShuffleNetV2 \cite{ma2018shufflenet}, ShuffleFaceNet models were proposed for face recognition, with parameter counts ranging from 0.5M to 4.5M and verification accuracies exceeding 99.20\% on the LFW dataset \cite{martindez2019shufflefacenet}. Neural architecture search (NAS) was utilized in \cite{boutros2022pocketnet} to automatically design an efficient network called PocketNet for face recognition. The PocketNet architecture was learned using the differential architecture search (DARTS) algorithm on the CASIA-WebFace dataset, and knowledge distillation (KD) was employed during training.  Yan et al. \cite{yan2019vargfacenet} employed knowledge distillation (KD) and variable group convolutions to address computational intensity imbalances in face recognition networks. Alansari et al. proposed GhostFaceNets, which exploit redundancy in convolutional layers to create compact networks \cite{alansari2023ghostfacenets}. These modules generate a fixed percentage of convolutional feature maps using computationally inexpensive depth-wise convolutions. Recently, George et al. introduced EdgeFace, a combination of CNN-Transformer architecture that achieved strong verification performance with minimal FLOP and parameter complexity \cite{george2023edgeface}.

\subsection{Knowledge Distillation}

The concept of Knowledge Distillation was first introduced by Hinton et al. \cite{hinton2015distilling}. The primary goal of knowledge distillation is to transfer the knowledge from a pre-trained, complex ``teacher" model to a simpler, more efficient ``student" model. The methods for distillation in classification tasks can primarily be learned through the utilization of soft labels from a teacher and ground truth \cite{hinton2015distilling}. Another approach involves feature-based learning, where the student aims to match the intermediate layers of the teacher \cite{romero2014fitnets}. Additionally, contrastive-based methods have also been employed \cite{tian2019contrastive} for distilling the knowledge of a teacher to a student.

Over the years, several methods have been proposed in the literature
\cite{romero2014fitnets,komodakis2017paying,kim2018paraphrasing,chen2021distilling,li2023curriculum,zhu2018knowledge,zhang2019your,chen2020online} to enhance the efficiency of distillation. However, most of these methods rely on the availability of original or similar training datasets, which can be limited due to security and privacy concerns. Consequently, traditional data-dependent distillation methods become impractical. To address this challenge, researchers have introduced Data-free knowledge distillation (DFKD), without relying on the original or real training data. DFKD aims to develop a distillation strategy using a synthesis-based approach. These approaches utilize either whitebox teacher model \cite{lopes2017data,chen2019data,yin2020dreaming} or data augmentation techniques \cite{asano2021augmented} to generate synthetic samples. These synthetic samples act as substitute training datasets for distillation. By training on such synthetic data, the student model can effectively learn from the teacher model without needing access to real training data making it privacy friendly.
Along the same lines, Boutros et al.~\cite{boutros2023unsupervised}
proposed an unsupervised face recognition model
based on unlabeled synthetic data. They used contrastive learning to maximize the similarity between two augmented images (using geometric and color transformations) of the same synthetic image. However, since the data augmentation cannot provide enough inter-class variations, it affects the performance of trained face recognition model when evaluating on benchmark datasets.

\section{Proposed Framework}\label{sec:proposed-method}
In this section, we describe our proposed framework for training a lightweight face recognition model using synthetic data using knowledge distillation. We describe the architecture of lightweight face recognition model in Section~\ref{subsec:proposed-method:network} and explain our  knowledge distillation framework using synthetic data in Section~\ref{subsec:proposed-method:distillation}.

\subsection{Lightweight Network Architecture}\label{subsec:proposed-method:network}
As discussed in Section~\ref{sec:related-works}, lightweight face recognition models in the literature usually adapt lightweight neural network models for face recognition tasks. However, our knowledge distillation framework can be applied to any lightweight model with only the condition that the output of the lightweight network should have the same dimensions as the embedding of the teacher model. To eliminate this condition so that the proposed framework can be used for any lightweight network with different output sizes, we use a fully connected layer at the output of the lightweight network to have output with the same size as the teacher model.

In this paper, we use TinyNet~\cite{han2020model} as the backbone for the lightweight FR model. 
The TinyNet is an optimized version of 
EfficientNet~\cite{tan2019efficientnet}, which  uses a structure that simultaneously enlarges the
resolution, depth, and width in a Rubik’s cube for neural networks and find networks with high efficiency by changing these three dimensions. However, authors in \cite{han2020model} show that the resolution and depth are more important than width for small networks, and propose smaller models derived from the EfficientNet-B0 as different variations of TinyNet, which are efficient and achieve high accuracy in recognition tasks.  
The feature layer of TinyNet has 1280 dimensions and the embedding of our teacher network has 512 dimensions. Therefore, we add a fully connected layer to generate 512-length feature at the output of TinyNet and call our lightweight face recognition network based on TinyNet \textit{TinyFaR}.
We should note that to our knowledge, TinyNet lightweight network structure has not been used before for face recognition in the literature.

\begin{figure}[t]
    \centering
    \includegraphics[width=0.99\linewidth]{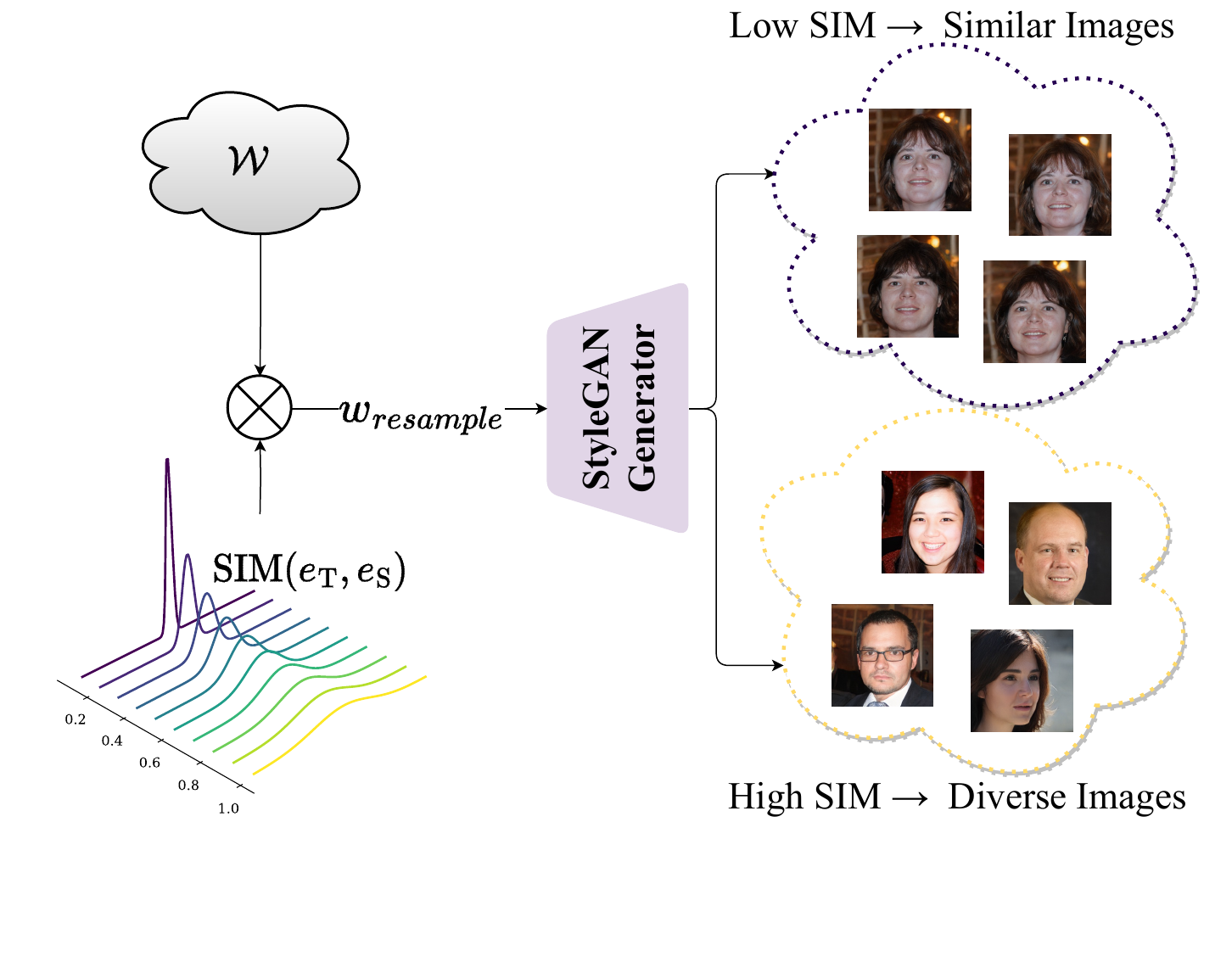}
    \caption{Schematic showing the re-sampling strategy in the proposed approach. When teacher-student agreement is high, the re-sampling method generates diverse images. Conversely, when the similarity is low, i.e, when the given sample is challenging, re-sampling generates similar (challenging) samples facilitating the learning.
    } \label{fig:resampling}
  \end{figure}

\subsection{Knowledge Distillation with Synthetic Data}\label{subsec:proposed-method:distillation}
Let $F_\text{T}$ and $F_\text{S}$ denote the teacher\footnote{Note that the teacher model can be blackbox and we do not use teacher's gradients in our method.} and student (lightweight) face recognition models, respectively.
In this paper, we consider StyleGAN~\cite{karras2020training} as a pretrained face generator model, which consists of a mapping network $M$ and a generator network $G$. The mapping network takes a a noise $\bm{z}\in\mathcal{Z}\sim N(0,\mathbb{I})$ from input latent space $\mathcal{Z}$ with Gaussian distribution and generates an intermediate latent code $\bm{w}\in\mathcal{W}$. Then, the intermediate latent code $\bm{w}$ is used by the generator network to generate a face image $I=G(\bm{w})$.
In our knowledge distillation framework, we first generate a batch of synthetic face images and extract the teacher's embeddings $e_\text{T}=F_\text{T}(I)$.
Then, we train the student network by minimizing the mean squared error (MSE) of the teacher and student's embeddings as follows:

\begin{equation} \label{eq:loss-KD}
	\mathcal{L}_\text{KD} =  \left\lVert \bm{e}_\text{T}- F_\text{S}(I) \right\rVert_2^2.
\end{equation}

Minimizing the MSE of embeddings helps the student network to extract embeddings similar to the teacher's embeddings from a given face image.

\begin{algorithm}[tb]
	\caption{Our proposed knowledge distillation approach}\label{alg:KD}
	\begin{algorithmic}[1]
		\Require   $n_{\text{epoch}}$: number of epochs, $n_{\text{iteration}}$:  number of iterations in each epoch, $\alpha$:  learning rate, $c$:  re-sampling coefficient.
		\Procedure{Training}{}
		\State Initialize weights $\theta_\text{S}$ of the student network
		\For{$\text{epoch} = 1, ..., n_{\text{epoch}}$}
		\For{$\text{itr} = 1, ..., n_{\text{iteration}}$}
            \State \textbf{Step 1}: Train with random samples
		\State \quad Sample a batch of  random noise vectors:
		\State \quad\quad $\bm{z}\in\mathcal{Z}\sim\mathcal{N}(0,\mathbb{I})$
		\State \quad Generate synthetic face images:
		\State \quad\quad $\bm{w}=M(\bm{z})$
            \State \quad\quad $\bm{I}=G(\bm{w})$
		\State \quad Extract teacher's embeddings $e_\text{T}$:
		\State \quad\quad $e_\text{T}=F_\text{T}(I)$
		\State \quad Calculate loss $\mathcal{L}_{\text{KD}}$ and optimize $\theta_\text{S}$:  
		\State \quad\quad $g_{\theta_\text{S}} \gets \nabla_{\theta_\text{S}} \mathcal{L}_{\text{KD}} $
		\State \quad \quad  ${\theta_\text{S}} \gets {\theta_\text{S}} - \alpha  \cdot \text{Adam}({\theta_\text{S}}, g_{\theta_\text{S}}) $
            \State \textbf{Step 2}: Train with dynamic re-sampling
		\State \quad Calculate similarity of $e_\text{T}$ and $e_\text{S}$:
            \State \quad \quad $\bm{s}_\text{sim} = \text{SIM}(e_\text{T}, e_\text{S})$
            \State \quad Re-sample based on similarity:
            \State \quad \quad $\bm{w}_\text{resample}=\bm{w}+c\times\bm{s}_\text{sim}\times \bm{n},$
            \State \quad \quad \quad where $\bm{n}\sim\mathcal{N}(0,\mathbb{I})$
		\State \quad Generate synthetic face images:
		\State \quad\quad $\bm{I}=G(\bm{w}_\text{resample})$
		\State \quad Extract teacher's embeddings $e_\text{T}$:
		\State \quad\quad $e_\text{T}=F_\text{T}(I)$
		\State \quad Calculate loss $\mathcal{L}_{\text{KD}}$ and optimize $\theta_\text{S}$:  
		\State \quad\quad $g_{\theta_\text{S}} \gets \nabla_{\theta_\text{S}} \mathcal{L}_{\text{KD}} $
		\State \quad \quad  ${\theta_\text{S}} \gets {\theta_\text{S}} - \alpha  \cdot \text{Adam}({\theta_\text{S}}, g_{\theta_\text{S}}) $
		\EndFor
		\EndFor
		\EndProcedure
	\end{algorithmic}
\end{algorithm}

After updating the weights of student network with our knowledge distillation loss $\mathcal{L}_\text{KD}$ (as in Eq.~\ref{eq:loss-KD}), we sample around the intermediate latent codes based on the similarity of embeddings extracted by the student $\bm{e}_\text{S}$ and teacher  $\bm{e}_\text{T}$ networks in our batch. To this end, we use the cosine similarity and normalize it in (0,1) interval as follows:
\begin{equation} \label{eq:sim}
	\text{SIM} (\bm{e}_\text{T}, \bm{e}_\text{S}) =  0.5\times(1+\frac{\bm{e}_\text{S} \cdot \bm{e}_\text{T}} { \left\lVert\bm{e}_\text{S}\right\rVert_2 \cdot \left\lVert \bm{e}_\text{T}\right\rVert_2}).
\end{equation}
Having normalized similarity score $\bm{s}_\text{sim} = \text{SIM}(e_\text{T}, e_\text{S})$, we re-sample around each latent code:
\begin{equation} \label{eq:resample}
\bm{w}_\text{resample}=\bm{w}+c\times\bm{s}_\text{sim}\times \bm{n},
\end{equation}
where $\bm{n}\sim\mathcal{N}(0,\mathbb{I})$ is a random noise with Gaussian distribution and $c$ is a constant coefficient. As a matter of fact, in our re-sampling based on similarity score $\bm{s}_\text{sim}$ as in Eq.~\ref{eq:resample}, we sample with higher standard deviation values around the latent codes which achieved higher similarity in our initial sampling,  and thus letting more variation in re-sampling. While, for the lower similarity between embeddings extracted by the student and teacher networks, the standard deviation values for re-sampling are smaller so that during re-sampling we can sample around the same latent codes. Therefore, our dynamic re-sampling approach helps us further sample difficult images while exploring the latent space. Fig.~\ref{fig:resampling} illustrates our re-sampling strategy.
After re-sampling new latent codes, we generate synthetic face images and optimize our student network with our knowledge distillation loss $\mathcal{L}_\text{KD}$ (as in Eq.~\ref{eq:loss-KD}).
Our knowledge distillation framework using synthetic data (named \textit{SynthDistill}) is depicted in Fig.~\ref{fig:framework} and summarized in Algorithm~\ref{alg:KD}.

\section{Experiments}\label{sec:experiments}
In this section, we report our experiments and discuss our results. First, in Section~\ref{subsec:exp:dataset}  we describe our evaluation datasets, and in Section~\ref{subsec:exp:training_details} we explain our training details. In Section~\ref{subsec:exp:comparison}, we compare our method with previous methods based on synthetic data for face recognition in the literature. Then, we report different ablation studies and discuss effect of each part in our proposed framework in Section~\ref{subsec:exp:ablatin}. 

\subsection{Datasets}\label{subsec:exp:dataset}

We evaluate our trained student models using five different benchmarking datasets. The datasets chosen for evaluation comprised Labeled Faces in the Wild (LFW) \cite{huang2008labeled}, Cross-age LFW (CA-LFW) \cite{zheng2017cross}, CrossPose LFW (CP-LFW) \cite{zheng2018cross}, Celebrities in Frontal-Profile in the Wild (CFP-FP) \cite{sengupta2016frontal}, AgeDB-30 \cite{moschoglou2017agedb}. To maintain consistency with previous work, we present recognition accuracy values on these datasets.

\begin{table}[tbh]
\centering
\caption{Complexity of different network structures}
\label{tab:tinymodels}
\resizebox{0.85\columnwidth}{!}{
\begin{tabular}{l|l|c|c}
\toprule
    Role in our KD & Network & M FLOPS & M Params \\ \midrule
\rowcolor{Gray}
Teacher & IResNet100& 24,179.2 & 65.2 \\ \midrule
\multirow{3}{*}{Student} &TinyFaR-A & 254.3                     & 5.6                        \\ 
& TinyFaR-B & 151.3                     & 3.1                        \\
& TinyFaR-C & 76.8                      & 1.8                     \\\bottomrule
\end{tabular}}
\end{table}

\subsection{Training Details}\label{subsec:exp:training_details}
For the teacher network, we use the pretrained ArcFace model\footnote{The performance of  our teacher network on our benchmarking datasets in terms of recognition accuracy  is as follows:
LFW	(99.77 ± 0.28),
CA-LFW	(96.10 ± 1.10),
CP-LFW	(92.88 ± 1.52),
CFP-FP	(96.27 ± 1.10), and
AgeDB-30	(98.25 ± 0.71).
} with IResnet100 backbone from Insightface~\cite{deng2019arcface} trained on the MS-Celeb dataset~\cite{guo2016ms}. 
The embedding of our teacher network has 512 dimensions, but the feature layer of TinyNet has 1280 dimensions. Therefore, as discussed in Section~\ref{sec:proposed-method} we use a fully connected layer at the output of our TinyNet model so that it can generate embeddings with the same dimension as the teacher's embeddings and call it \textit{TinyFaR}.
In our experiments, we use different variations of TinyNet~\cite{han2020model} and build corresponding version of TinyFaR with 512-length feature as our student (lightweight) network. 
Table~\ref{tab:tinymodels} compares IResnet100 with different variations of TinyFaR in terms of computation complexity and number of parameters.  
We use StyleGAN2-ADA model \cite{karras2020training} to generate synthetic face images with $256\times256$ resolution and crop and resize images to have $112\times112$ face images for our knowledge distillation. We train our student networks with 17 epochs, where in each epoch we sampled one million images in step 1 of our algorithm~\ref{alg:KD} and re-sampled the same number of images with the re-sampling coefficient of $c=1$. We trained our student networks using Adam optimizer~\cite{kingma2014adam} on a system equipped with a single NVIDIA GeForce RTX\textsuperscript{TM} 3090. 
For training face recognition from scratch in our experiments, we used CosFace \cite{wang2018cosface} loss function.
The source codes of our experiments are publicly available\footnote{Source code: \href{https://gitlab.idiap.ch/bob/bob.paper.ijcb2023_synthdistill}{https://gitlab.idiap.ch/bob/bob.paper.ijcb2023\_synthdistill}}.

\begin{table}[tb]
\centering
\caption{Synthetic and real face datasets}
\label{tab:datasets}
\resizebox{1\columnwidth}{!}{
\begin{tabular}{l|c|c|c|l}
\toprule
Dataset   & \#Images  &\#Subjects &Data      & Method   \\ \midrule
\rowcolor{Gray}
WebFace-4M \cite{zhu2021webface260m} & 4,235,242  & 205,990   & Real         & Web-crawled          \\ 
SFace \cite{boutros2022sface} (IJCB 2022)    & 1,885,877  & 10,572    & Synthetic          & StyleGAN model         \\ 
DigiFace \cite{bae2023digiface} (WACV 2023)  & 1,219,995  & 109,999   & Synthetic         & Rendering          \\ 
DCFace \cite{kim2023dcface}  (CVPR 2023)  & 1,300,000  & 60,000    & Synthetic        & Diffusion model         \\ \bottomrule
\end{tabular}}
\end{table}

\begin{table*}[]
\centering
\renewcommand{\arraystretch}{1.15}
\caption{Comparison of our knowledge distillation approach with training from scratch using other synthetic datasets}\label{tab:comparison}
\resizebox{0.95\textwidth}{!}{
\begin{tabular}{cccccccc}
\toprule
Network &   Training   &  Dataset &           LFW &         CA-LFW &         CP-LFW &        CFP-FP &      AgeDB-30 \\
\midrule
\rowcolor{Gray}
\cellcolor{white}
\multirow{5}{*}{TinyFaR-A} &   \cellcolor{white}\multirow{4}{*}{Classification} &     WebFace-4M (real) &   99.58 ± 0.37 &  95.02 ± 1.00 &  91.82 ± 1.29 &  95.09 ± 1.15 &  94.62 ± 1.21 \\
& &  DCFace (synthetic)  &   97.35 ± 0.66 &  90.08 ± 1.27 &  79.63 ± 2.08 &  82.01 ± 1.62 &  85.12 ± 2.05 \\
 & &    SFace (synthetic) &   90.48 ± 1.54 &  75.48 ± 2.27 &  71.40 ± 1.89 &  72.07 ± 2.38 &  68.65 ± 2.53 \\
 & & DigiFace (synthetic) &   89.12 ± 1.30 &  71.65 ± 2.14 &  69.63 ± 1.70 &  76.24 ± 1.34 &  68.60 ± 1.23 \\\cline{2-8}
& Knowledge Distillation &\textbf{SynthDistill (synthetic) [ours]} & \textbf{99.52 ± 0.31} & \textbf{94.57 ± 1.01} & \textbf{87.00 ± 1.64} & \textbf{90.89 ± 1.54} & \textbf{94.93 ± 1.35}\\
\midrule
\rowcolor{Gray}
\cellcolor{white}
\multirow{5}{*}{TinyFaR-B} &   \cellcolor{white}\multirow{4}{*}{Classification} &    WebFace-4M (real) &   99.55 ± 0.40 &  94.73 ± 0.88 &  90.95 ± 1.43 &  94.00 ± 1.23 &  93.72 ± 1.37 \\
 &  & DCFace (synthetic) &   97.40 ± 0.75 &  89.62 ± 1.37 &  78.93 ± 1.74 &  82.47 ± 1.74 &  85.03 ± 1.97 \\
 &   &  SFace (synthetic) &   91.10 ± 1.22 &  76.15 ± 1.46 &  72.02 ± 1.34 &  71.13 ± 2.43 &  68.73 ± 1.68 \\
 &  &DigiFace (synthetic) &   88.03 ± 1.05 &  70.27 ± 2.17 &  68.22 ± 1.74 &  75.29 ± 2.14 &  66.38 ± 1.82 \\\cline{2-8}
& Knowledge Distillation &\textbf{SynthDistill (synthetic) [ours]}  &  \textbf{99.20 ± 0.41} & \textbf{93.78 ± 0.78} & \textbf{84.93 ± 2.10} & \textbf{88.19 ± 1.34} & \textbf{93.02 ± 1.30}\\
\midrule
\rowcolor{Gray}
\cellcolor{white}
\multirow{5}{*}{TinyFaR-C} &   \cellcolor{white}\multirow{4}{*}{Classification} &    WebFace-4M (real) &   99.37 ± 0.26 &  93.08 ± 1.11 &  88.98 ± 1.12 &  92.30 ± 1.74 &  91.18 ± 1.80 \\
 & &  DCFace (synthetic) &   96.78 ± 0.73 &  88.48 ± 1.02 &  77.22 ± 1.80 &  80.59 ± 1.80 &  83.65 ± 2.14 \\
 &  &   SFace (synthetic) &   91.12 ± 1.01 &  76.70 ± 1.25 &  71.27 ± 1.98 &  72.24 ± 1.53 &  71.13 ± 1.35 \\
 & & DigiFace (synthetic) &   87.47 ± 0.87 &  69.18 ± 1.94 &  68.05 ± 1.94 &  74.16 ± 2.68 &  67.23 ± 1.85 \\\cline{2-8}
& Knowledge Distillation &\textbf{SynthDistill (synthetic) [ours]}  & \textbf{98.58 ± 0.44} & \textbf{91.80 ± 1.04} & \textbf{82.00 ± 2.14} & \textbf{84.54 ± 1.57} & \textbf{88.98 ± 1.49}\\
\bottomrule
\end{tabular}}
\end{table*}

\begin{table*}[tbh]
\centering
\caption{Ablation study on the effect of dynamic sampling}\label{tab:ablation:dynamic_sampling}
\resizebox{0.8\textwidth}{!}{

\begin{tabular}{@{}lcccccc@{}}
\toprule
Sampling & \# Samples/epoch      & LFW &         CA-LFW &         CP-LFW &        CFP-FP &      AgeDB-30    \\ \midrule
static & 1 M   &  98.87 ± 0.39 &  92.93 ± 0.94 &  83.52 ± 1.63 &  87.60 ± 1.44 &  91.25 ± 2.18  \\
static & 2 M   &  98.95 ± 0.47 &  93.67 ± 0.78 &  84.75 ± 1.94 &  88.51 ± 1.63 &  92.83 ± 1.76  \\
dynamic (re-sampling in $\mathcal{Z}$)& 1M + 1M  & 99.28 ± 0.30 & 93.88 ± 1.09 & 84.45 ± 1.95 & 87.59 ± 1.20 & 92.45 ± 1.69 \\ 
dynamic (re-sampling in $\mathcal{W}$)& 1M + 1M   & 99.52 ± 0.31 & 94.57 ± 1.01 & 87.00 ± 1.64 & 90.89 ± 1.54 & 94.93 ± 1.35 \\\bottomrule
\end{tabular}}
\end{table*}

\subsection{Comparison}\label{subsec:exp:comparison}
We compare the performance of our proposed knowledge distillation framework with training the same network using synthetic datasets in the literature, including DigiFace~\cite{bae2023digiface},  SFace~\cite{boutros2022sface}, and DCFace~\cite{kim2023dcface}. In addition, we also consider training with real data using WebFace-4M \cite{zhu2021webface260m} as our baseline.  
Table~\ref{tab:datasets} compares these datasets in terms of the number of images and samples and their generation method. All these datasets are generated to have inter-class and intra-class variation, and thus have identity labels. Therefore, these datasets can be used for training lightweight face recognition from scratch  using the classification training. In contrast, our proposed framework based on dynamic sampling approach does not provide idenity labels and can be used within a knowledge distillation training. 
Table~\ref{tab:comparison} reports the recognition performance of different variations of TinyFaR when training with datasets. 
As the results in this table show, our knowledge distillation approach with synthetic data (and no identity labels) far outperforms training from scratch using synthetic data and has comparable performance with training using real data.

\subsection{Ablation studies}\label{subsec:exp:ablatin}
\paragraph{Effect of dynamic sampling:}
To evaluate the effect of dynamic sampling in our proposed framework, we compare the performance network trained with knowledge distillation using our dynamic sampling (sampling + re-sampling) using static sampling (with no re-sampling). Table~\ref{tab:ablation:dynamic_sampling} compares the performance of TinyFaR-A trained with knowledge distillation using our dynamic sampling (sampling + re-sampling in $\mathcal{W}$ space) with one million samples plus one million re-sampling (1M+1M) in each epoch as well as static sampling with one million and two million samples in each epoch. As the results in this table show knowledge distillation using our dynamic sampling with one million iterations in each epoch outperforms the same number of iterations or sample total samples with static sampling. This table also compares our dynamic re-sampling in $\mathcal{W}$ space to dynamic re-sampling in $\mathcal{Z}$ space.
As the results show dynamic re-sampling in both spaces achieves better performance than static sampling. In addition,  comparing dynamic re-sampling space, the results show that dynamic re-sampling in $\mathcal{W}$ leads to superior performance.

\begin{table}[tbh]
\centering
\setlength{\tabcolsep}{4pt}
\caption{Ablation study on the effect of number of  sampling}\label{tab:ablation:number_sampling}
\resizebox{1\columnwidth}{!}{

\begin{tabular}{@{}lcccccc@{}}
\toprule
\# Itr  & LFW &         CA-LFW &         CP-LFW &        CFP-FP &      AgeDB-30    \\ \midrule
0.5 M   &  99.43 ± 0.37 &  93.90 ± 1.11 &  86.13 ± 1.81 &  89.46 ± 1.48 &  93.53 ± 1.36  \\
1 M   & 99.52 ± 0.31 & 94.57 ± 1.01 & 87.00 ± 1.64 & 90.89 ± 1.54 & 94.93 ± 1.35\\
2 M   & 99.48 ± 0.39 &  95.07 ± 0.97 &  87.78 ± 1.64 &  91.31 ± 1.99 &  95.25 ± 1.19  \\\bottomrule
\end{tabular}}
\end{table}

\paragraph{Effect of number of  sampled images:}
To evaluate the effect of the number of sample images in our dynamic sampling, we train TinyFaR-A with different numbers of iterations (sampling and re-sampling) per epoch in our knowledge distillation approach. Table~\ref{tab:ablation:number_sampling} reports the performance of the trained model with different numbers of iterations. As the results in this table show, higher iterations help our knowledge distillation with the cost of more training computation. However, to reduce computations in our experiments we use one million iterations (1M sampling + 1M re-sampling) in our experiments.

\begin{table}[tbh]
\centering
\setlength{\tabcolsep}{4pt}
\caption{Ablation study on the effect of re-sampling coefficient}\label{tab:ablation:number_sampling}
\resizebox{1\columnwidth}{!}{

\begin{tabular}{@{}ccccccc@{}}
\toprule
coef. ($c$)  & LFW &         CA-LFW &         CP-LFW &        CFP-FP &      AgeDB-30    \\ \midrule
0.8   &  99.45 ± 0.32 & 94.58 ± 0.95 & 86.10 ± 2.23 & 90.23 ± 1.68 & 94.82 ± 1.15 \\
0.9   &  99.40 ± 0.41 & 94.90 ± 1.09 & 87.23 ± 2.02 & 90.36 ± 1.45 & 94.72 ± 1.07 \\
1   & 99.52 ± 0.31 & 94.57 ± 1.01 & 87.00 ± 1.64 & 90.89 ± 1.54 & 94.93 ± 1.35\\
1.1   &  99.47 ± 0.44 & 94.95 ± 0.84 & 87.53 ± 1.78 & 90.81 ± 1.61 & 95.13 ± 1.08 \\
1.2   &  99.53 ± 0.32 & 94.95 ± 0.90 & 87.47 ± 1.27 & 90.94 ± 1.63 & 94.52 ± 1.47\\
1.3    & 99.52 ± 0.31 & 94.50 ± 0.97 & 87.58 ± 1.84 & 91.17 ± 1.50 & 95.05 ± 1.28 \\
1.4   & 99.48 ± 0.32 & 94.77 ± 0.97 & 87.40 ± 1.74 & 90.56 ± 1.49 & 94.78 ± 1.29 \\
1.5   & 99.47 ± 0.32 & 94.58 ± 1.00 & 88.17 ± 1.64 & 90.84 ± 1.24 & 94.80 ± 1.07 \\
\bottomrule
\end{tabular}}
\end{table}

\paragraph{Effect of re-sampling coefficient:}
As another ablation study, we evaluate the effect of re-sampling coefficient $c$ in our dynamic sampling. Table~\ref{tab:ablation:number_sampling} reports the performance of TinyFaR-A trained with our knowledge distillation using different re-sampling coefficient values. As the results in this table show, with a higher re-sampling coefficient our dynamic re-sampling can generate more diverse images and achieve higher recognition performance. However, a very high re-sampling coefficient can also cause $\bm{w}_\text{resample}$ to be out of the distribution of $\mathcal{W}$, and thus drop the performance.

\section{Discussions}\label{sec:discusssion}
The results in Table~\ref{tab:comparison} show that our proposed knowledge distillation framework outperforms training using synthetic datasets in the literature and achieves comparable performance with training using real face images. Comparing the performance of networks trained with previous synthetic datasets to networks trained with real data, we observe a considerable gap in the performance of trained face recognition models with synthetic and real data. Meanwhile, our proposed knowledge distillation method still achieves lower but is very close to the performance of training with real data. 

Unlike previous synthetic face datasets, our method does not require identity labels, and thus does not have many issues in generating synthetic datasets with inter-class and intra-class variations. Instead, our knowledge distillation approach with dynamic sampling leverages the most capacity of StyleGAN to generate training samples, which helps to achieve comparable performance to training with real data. 
Our proposed framework avoids the requirements of hard identity labels for the generated images, which further assists the generation network to produce challenging samples though a feedback mechanism during our knowledge distillation, thus enabling the training of much robust models.
We should also note that, for generation of synthetic face datsets in the literature, a pretrained face recognition model (which has been trained on a large-scale real face recognition dataset) is used in the process of generation of synthetic dataset. Therefore, training with synthetic face datasets in the literature indirectly benefits from the information and knowledge of the pretrained face recognition model (trained on real images) used for generating the synthetic dataset. 
In our proposed framework,  we also use the pretrained face recognition model, but instead of following common two-step approach (generation of dataset and training with new dataset), we use the pretrained face recognition model as a teacher in our knowledge distillation approach and generate synthetic face images used in our training with no identity label.

Our ablation studies show the effect of each part in our knowledge distillation framework. In particular, the results demonstrate that our dynamic sampling improves our knowledge distillation compared to static sampling. In addition, using our dynamic sampling and with more number of iterations or higher re-sampling coefficient can improve the knowledge distillation, as it helps our student to learn embeddings of more face images from the teacher.  

\section{Conclusions}\label{sec:conclusion}
In this paper, we proposed a data-free framework (named \textit{SynthDistill}) to  train lightweight face recognition models based on knowledge distillation using synthetic data. 
We combined the two steps of  data generation and training the lightweight network and have an online-generation and training in the loop using a distillation framework. 
We dynamically generated synthetic face images during training and distilled the knowledge of a pretrained and blackbox face recognition model. Our dynamic sampling helps our student network to further see difficult samples while exploring new samples, leading to more robust training. Our knowledge distillation framework does not require identity-labeled training data, and thus mitigates challenges in generating intra-class variations in synthesized datasets. 
We adapted the TinyNet architecture to use in our knowledge distillation framework and trained lightweight face recognition models (called \textit{TinyFaR}).  
We reported extensive experimental evaluation on five different face recognition benchmarking datasets, 
including LFW, CA-LFW, CP-LFW, CFP-FP, and AgeDB-30. The experimental results demonstrate the superiority of our proposed knowledge distillation approach compared to training previous synthetic datasets. 

Our experimental results also showed that while there is a considerable gap between training with synthetic datasets and real data, our knowledge distillation framework based on synthetic data achieves comparable performance with training with real data and significantly reduces the gap between  models trained on synthetic data and models trained on real data. Achieving such an improvement in training using synthetic data within our proposed framework shows more potential in training with synthetic data and motivates further research on training with synthetic data.
Furthermore, our results for  lightweight student networks pave the way for developing privacy-aware and resource-efficient face recognition models.

\section*{Acknowledgments}
This research is based upon work supported by the H2020 TReSPAsS-ETN Marie Sk\l{}odowska-Curie early training network (grant agreement 860813). 

This research is also based upon work supported in part by the Office of the Director of National Intelligence (ODNI), Intelligence Advanced Research Projects Activity (IARPA), via [2022-21102100007]. The views and conclusions contained herein are those of the authors and should not be interpreted as necessarily representing the official policies, either expressed or implied, of ODNI, IARPA, or the U.S. Government. The U.S. Government is authorized to reproduce and distribute reprints for governmental purposes notwithstanding
 any copyright annotation therein.

{\small
\bibliographystyle{ieee}
\bibliography{egbib}
}

\end{document}